\documentclass[a4paper, 11pt, oneside]{article}

\usepackage[english]{babel}
\usepackage[utf8]{inputenc}
\usepackage{hyperref}
\usepackage{amsmath}
\usepackage{graphicx}
\usepackage{booktabs}
\usepackage{amssymb}
\usepackage{subfig}

\usepackage{tikz}
\usepackage{pgfplots}
\usepackage{graphicx,verbatimbox}
\usetikzlibrary{positioning}
\usetikzlibrary{intersections}
\usetikzlibrary{calc}
\usepgfplotslibrary{fillbetween}

\usepackage[
	left = 25mm,
	right = 25mm,
	top = 25mm,
	bottom = 25mm
]{geometry} 
\usepackage[font=footnotesize,labelfont=bf]{caption}

\providecommand{\keywords}[1]
{
  \small	
  \textbf{Keywords:} #1
}

\newcommand*\samethanks[1][\value{footnote}]{\footnotemark[#1]}

\interfootnotelinepenalty=\@M

\title{Predicting the Popularity of Games on Steam}

\author{
    Andra\v{z} De Luisa
	\thanks{Faculty of Computer and Information Science, University of Ljubljana}
	\and
	Jan Hartman
	\samethanks
	\and
	David Nabergoj
	\samethanks
	\and
	Samo Pahor
	\samethanks
	\and
	Marko Rus
	\samethanks
	\and
	Bozhidar Stevanoski
	\samethanks
    \and 
    Jure Dem\v{s}ar
    \samethanks
    \and
    Erik \v{S}trumbelj
    \samethanks
}

\date{}

\begin{document}
\maketitle
\thispagestyle{empty}

\begin{abstract}             
The video game industry has seen rapid growth over the last decade. Thousands of video games are released and played by millions of people every year, creating a large community of players. Steam is a leading gaming platform and social networking site, which allows its users to purchase and store games. A by-product of Steam is a large database of information about games, players, and gaming behavior. In this paper, we take recent video games released on Steam and aim to discover the relation between game popularity and a game's features that can be acquired through Steam. We approach this task by predicting the popularity of Steam games in the early stages after their release and we use a Bayesian approach to understand the influence of a game's price, size, supported languages, release date, and genres on its player count. We implement several models and discover that a genre-based hierarchical approach achieves the best performance. We further analyze the model and interpret its coefficients, which indicate that games released at the beginning of the month and games of certain genres correlate with game popularity.
\end{abstract}


\section{Introduction}
Steam is a video game digital distribution service owned by the Valve Corporation. It is currently the most widely used video game platform on personal computers, having published more than 8000 games in 2019 alone.\footnote{Source: \url{steamspy.com}, accessed May 1, 2021.}
Some of these games instantly reached a large community of players, while many others remained unpopular. It is difficult to fully understand how a game's popularity changes after its release, as this is influenced by many factors which are often difficult to measure or quantify. For example, marketing and budget information may be of great value for such a model but are generally difficult to obtain. Furthermore, popularity heavily relies on a great player experience, which is influenced by many factors, such as the complexity of the game's story, graphics, player's interaction, etc. These attributes can only be measured through some kind of operationalization, for instance by analyzing reviews or observing the number of Google searches for a game. 

However, basic properties of the game may also influence its popularity. 
In this study, we focus on discovering how such properties may lead to improved chances for the successful release of the game. 
For instance, we observe a game's price and reason whether it is more sensible to release a game for free and reach many players, or release it at a higher cost, which may indicate that the game is of higher quality.
We also address questions like how popularity differs among different genres and what genre should a game have to most likely succeed.
Any kind of such information can help video game developers make design decisions about their games.
We approach this problem from a Bayesian perspective, as it allows us to compute robust uncertainty estimates that may be highly relevant for making potential business decisions.
For example, if the model predicts the player count with high certainty, the developer may confidently incorporate changes to the game development process.
However, if the model's uncertainty is high, then it may be better to further analyze the state of development.

In the following subsection, we briefly review some related work on analyzing the success of Steam games.
In Section~\ref{sec:dataset}, we describe the procedure of obtaining Steam game data and what preprocessing steps were taken to improve its quality.
In Section~\ref{sec:methodology}, we thoroughly describe how the data was transformed to be used in our models.
Later on in the section, we also describe the models and the motivation behind them.
We list the quantitative results in Section~\ref{sec:results}. We also provide visualizations regarding model performance and effects that different features have on generated predictions.
Finally, we review the key points of the paper in Section~\ref{sec:conclusion}, where we also give some directions for future work.

\subsection{Related Work}
Game popularity can be defined and measured in numerous ways. The simplest and most intuitive operationalization of game popularity is the number of players in a set period of time since a game is played by more players if more people know about it. Budiarto et al.~\cite{budiarto2018game} also use four other metrics and combine them to calculate game popularity: user count, unique page views, average time on the page, the difference of unique page views, and average time on the page from the day before. They gather this information using data from Google Analytics platform, which is updated daily. They also emphasize that player count is the primary explanatory variate for understanding game popularity, which is why we use it as the only target variable.

Ahn et al.~\cite{ahn2017makes} try to characterize popular and unpopular games by observing game reviews obtained from Steam. They identify multiple types of reviews that are associated with the popularity of a game. This reveals what users like and what makes them dislike a game. If certain reviews associated with low popularity appear on Steam, one can then take action and try to improve the game. Lin et al.~\cite{lin2019empirical} also close the gap between the domain of game reviews and the domain of app reviews, and provide insights into when negative reviews are most likely to be posted. This may help understand player satisfaction with the game and help explain the popularity of the game, but since not many reviews are posted for unpopular games and because of possible bias or misleading reviews, we decided not to use them in our analysis.

Sometimes we are only concerned about a specific property of a video game and want to explore how it affects its popularity. Lin et al. \cite{lin2018empirical} analyze the advantages of \textit{early access} games, where players can purchase and play a game before its official release. Authors observe that in the early access stage, the average rating of the reviews is much higher than later, suggesting that players are more tolerant of imperfections in the early access stage. Therefore, it is reasonable to make the game accessible early, as positive reviews may attract more players later. 
We include this idea in our analysis and generalize it to consider not only the \textit{early access} tag, but also other game tags (mainly genres).

\section{Dataset}\label{sec:dataset}
We used several data sources to collect diverse information about Steam games. To capture the most recent state of the video gaming market, we only focus on recent games, released after 2015.

In the next sections, we list the sources of our collected data and provide a short summary of how the data was collected and processed.
\subsection{Data Collection}
The data was acquired from several sources: Steam, SteamSpy~\cite{steamspysite}, and SteamDB~\cite{steamdbsite}. 

SteamSpy and SteamDB are independent sources unaffiliated with Steam itself, but offer data that cannot be acquired through Steam, e.g.\ SteamDB collects and stores the past numbers of concurrent players, while Steam only offers the current number of concurrent players. We scraped the data over the course of one week in mid December 2020 using the relevant APIs for each of the data sources. To make sure that our scraper functioned correctly, we manually verified the scraped information on a small random sample by checking if what we retrieved matched the data displayed on the websites.

Steam's official API allows us to access the list of all items available in the Steam store. However, in addition to games, the list also includes various other items, like additional downloadable content or even non-game software. The API does not provide an easy way to filter such items. To address this issue, we used SteamSpy's listing of standalone games. After obtaining the list of standalone games, we used Steam's API to fetch all relevant information for each individual game. The data we fetched for each game included the list of genres, developers and publishers, the game's price, release date, number of supported languages, and description of system requirements. The final step of data collection was fetching the history of concurrent player counts from SteamDB. Here, we fetched the daily player count history for all games, released after 2015. 

\subsection{Data Preprocessing}
To make the collected data usable for analysis and modeling, we performed data preprocessing and cleaning. The first transformations were related to price and system requirements. Steam provides the game price in different currencies. We converted all prices to Euro and applied static conversion rates: 1 USD = 0.82 EUR and 1 GBP = 1.09 EUR (rates in November 2020). System requirements were retrieved as unstructured text data. We extracted the storage requirements from text data and converted them to MB. 
We performed a manual inspection and removed a selection of games with unreasonably high storage requirements (e.g. \textit{HuH?: and the Adventures of something}, which requires 9000 GB of storage). Our experiments were performed on a dataset of 8000 games.

\section{Methodology}\label{sec:methodology}

In this section, we illustrate our feature engineering process, further elaborate on the collected raw features as well as give additional insights and statistical information about both raw and engineered features. Afterwards, we present our predictive models and provide justifications for our choices of priors.

\subsection{Feature Engineering}
We have operationalized response variate as the median player count in the second month since a game's release; this was regarded as the game's future player count. We considered a game's median player count in the first month since release as the main explanatory variate; analogously this was regarded as the game's past player count. In addition, we also constructed features by considering specific information about the game.

Contrary to past player count, which is a reflection of the game's performance after it has already been released, some information is known at the moment of the release and plays a role when a player is deciding to buy a game. For instance, player might opt not to buy a certain game due to its price point. The game properties we used as features were game price, the number of supported languages, and game storage requirements.

\begin{figure}
    \centering
    \includegraphics[width=\hsize]{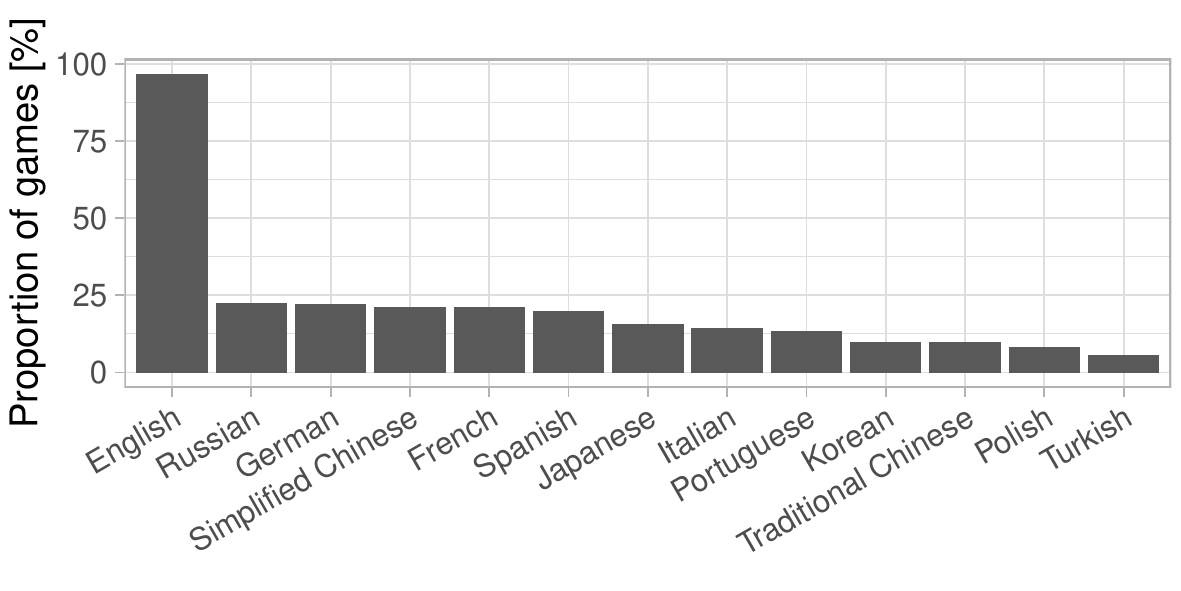}
    \caption{
    Support of the most popular languages, we can see that English is by far the most supported language. Note that one game can support more languages.
    }
    \label{fig:languages}
\end{figure}

There exist different motivations for including game price as an explanatory variate. The simplest one is that cheaper games are generally more popular since more people can afford them. This was already suggested in research related to price dynamics of video game consoles~\cite{liu_2010}. Policy simulations concluded that Nintendo could have won the commercial sales competition with Sony by reducing the cost of its products, thus reaching a larger user base. Our justification for including the number of supported languages was similar. We suspected that games with a larger number of supported languages would reach a larger player base. We show the distribution of the top supported languages in Figure~\ref{fig:languages}.

Game storage requirements were another included feature. A whitepaper article published by Limelight Networks \cite{limenet_2020} reinforces the decision of including this feature. Data based on worldwide survey results reported that 87 percent of players find the process of downloading games frustrating. Larger games could therefore suffer in popularity due to their longer download times as well as the longer download times of their potential updates. Conversely, a larger storage size could indicate that a game has more playable content, resulting in longer player engagement. Storage size could also indicate the general scale and budget of the game, since larger games tend to come from more prestigious studios.

Lastly, we constructed different temporal features from the release date timestamp.
For each game, we extracted the \textit{release year} value and information about the actual release day, both in a yearly as well as monthly context. 
We denoted these features as \textit{release day} and \textit{monthly release day}, respectively. 
To illustrate, if a game was released on March 2, 2020, it would have a release year value of 2020, a release day value of 62, since it was released on the sixty-second day of the year, and a monthly release day value of 2, since it was released on the second day of the month. 
The motivation behind these features was to capture different seasonalities present in player behavior or particular to the Steam platform or to capture the effect of different commercial years on game popularity. 
Concretely, release day (an integer between $1$ and $366$) was included to capture yearly seasonality related to possible annual online sales, holiday purchases, an increase in gaming during specific seasons, etc.
Past studies observed a statistically significant impact of seasonality on monthly playtime~\cite{palomba_2019}.
Monthly release day (an integer between $1$ and $31$) was included to capture monthly seasonality, such as the inclination to purchase or play games after receiving a monthly paycheck.

\subsection{Data Insights}
In this section, we present a few insights into the data which are important for explaining our decisions related to modeling. We cover the variates present in the data, showcase some of their properties and try to find correlations between them and the response variate. 

\subsubsection{Basic Numeric Variates}
The three basic numeric variates are the number of supported languages, the storage requirements, and the price. We show a few statistics in Table~\ref{tab:basicvariates}. All of them are heavily skewed towards 0, which makes sense -- most games will not have support for many languages, most games are not that large, and most games are not particularly expensive, or are actually free. We can also observe outliers present in all variates by noting their high standard deviations. This is particularly notable in the case of storage requirements. 

\begin{table}[htbp]
    \caption{Numeric variates statistics.}
    \centering
    \begin{tabular}{lrrrrr}
        \toprule
        name & mean & median & stdev & min & max \\
        \midrule
        number of languages & 4.800 & 2.000 & 5.500 & 1.000 & 29.000 \\
        storage requirements (GB) & 4.600 & 1.000 & 10.100 & 0.001 & 256.000 \\
        price (EUR) & 9.920 & 6.750 & 12.390 & 0.000 & 325.910 \\
        \bottomrule
    \end{tabular}
    \label{tab:basicvariates}
\end{table}

\subsubsection{Genres, Developers, and Publishers}
In addition to numeric variates, we also deal with the following categorical variates: a game's genres,\footnote{We follow Steam's broader definition of the word genre, which also includes information such as early access.} its developers, and its publishers. These variables have a common property -- a game can have more than one of them and vice versa (i.e.\ a many-to-many relationship). Due to this, they require transformations to be usable in our models. 

In our dataset, there are 33 genres, over 19000 developers, and over 23000 publishers. The large number of developers and publishers relative to the number of games is not a good sign for using them as features in our model. This means that most of them have likely not made many games and thus it is very hard to learn anything about them. In fact, over 15000 developers and 18000 publishers have only made one game. Over 18000 developers and 22000 publishers have made fewer than five games, leaving us with very few of those who have made more games and will bring a benefit if they are added into our models. Another possibility could be to group the small developers or publishers, but this would bring significant confusion in the model, and given that the cardinality would still remain fairly high, we decided to discard them from the modeling process.

\begin{figure}
    \centering
    \includegraphics[width=0.9\hsize]{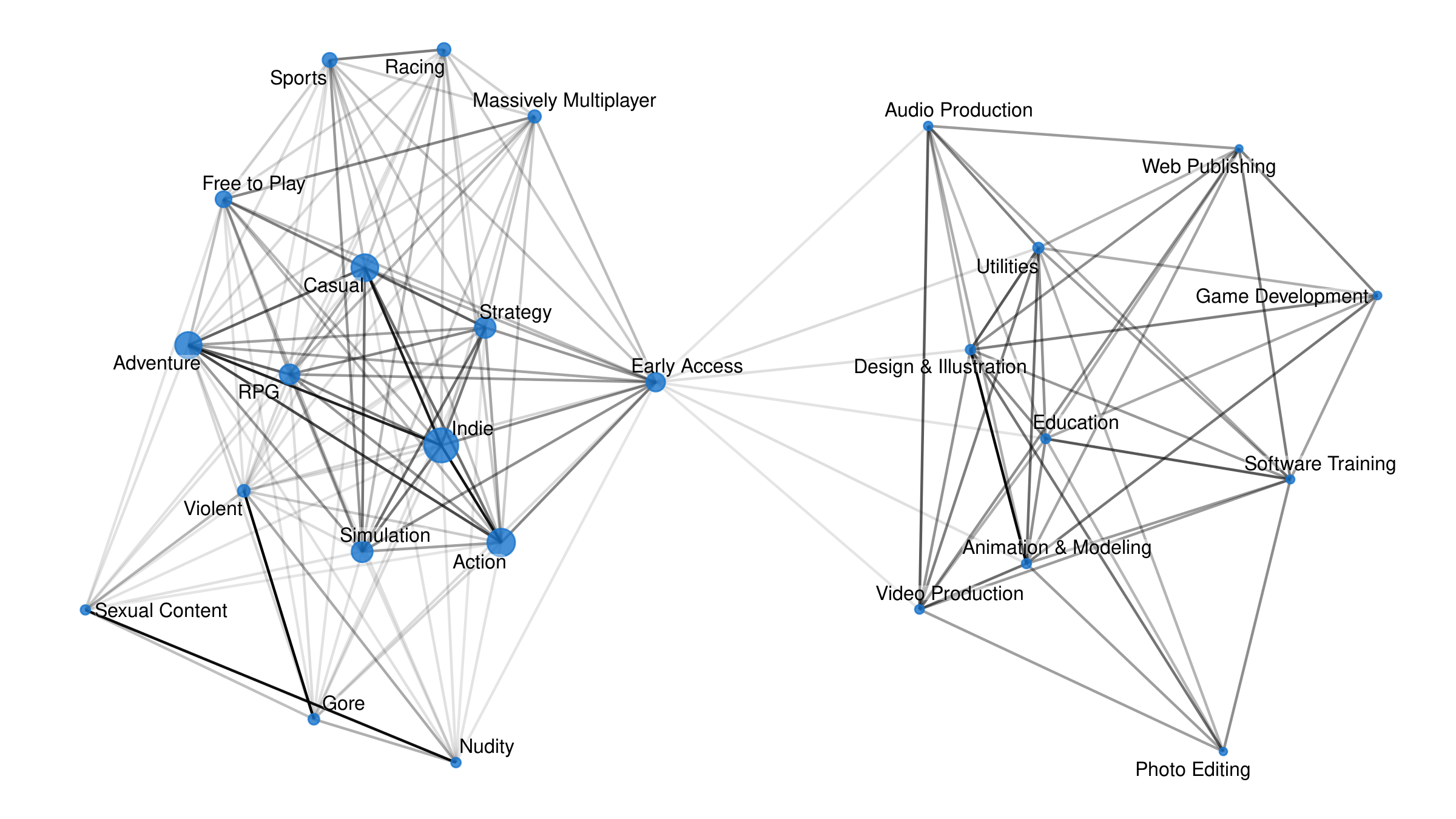}
    \caption{
        \textbf{Network plot of game genres}.
        The nodes represent genres while the edges represent games that are in both genres. The opacity of an edge represents the proportion of games the connected genres share. The size of a node represents the number of games in a genre.
    }
    \label{fig:genrenetwork}
\end{figure}

Conversely, the genres are more useful. The distribution of the numbers of games belonging to genres is still skewed since there are many genres with a small number of games, but not as extremely as in the cases of developers and publishers. There are also only 33 genres, so cardinality is not a problem and utilizing genres in our models is thus much easier. To see how the different genres connect, we visualize the counts of games and the connections between genres in \figurename~\ref{fig:genrenetwork}. We observe that indie, action, casual, and adventure games are the most common. Interestingly, two clusters emerge -- actual games and game-related utilities. They have only one common genre -- early access. From the opacity of the edges (i.e. the proportion of games the genres share relative to the total number of games that belong to those genres), we can see which genres have more in common. 

\subsubsection{Player Counts}

\begin{figure}
    \centering
    \includegraphics[width=\hsize]{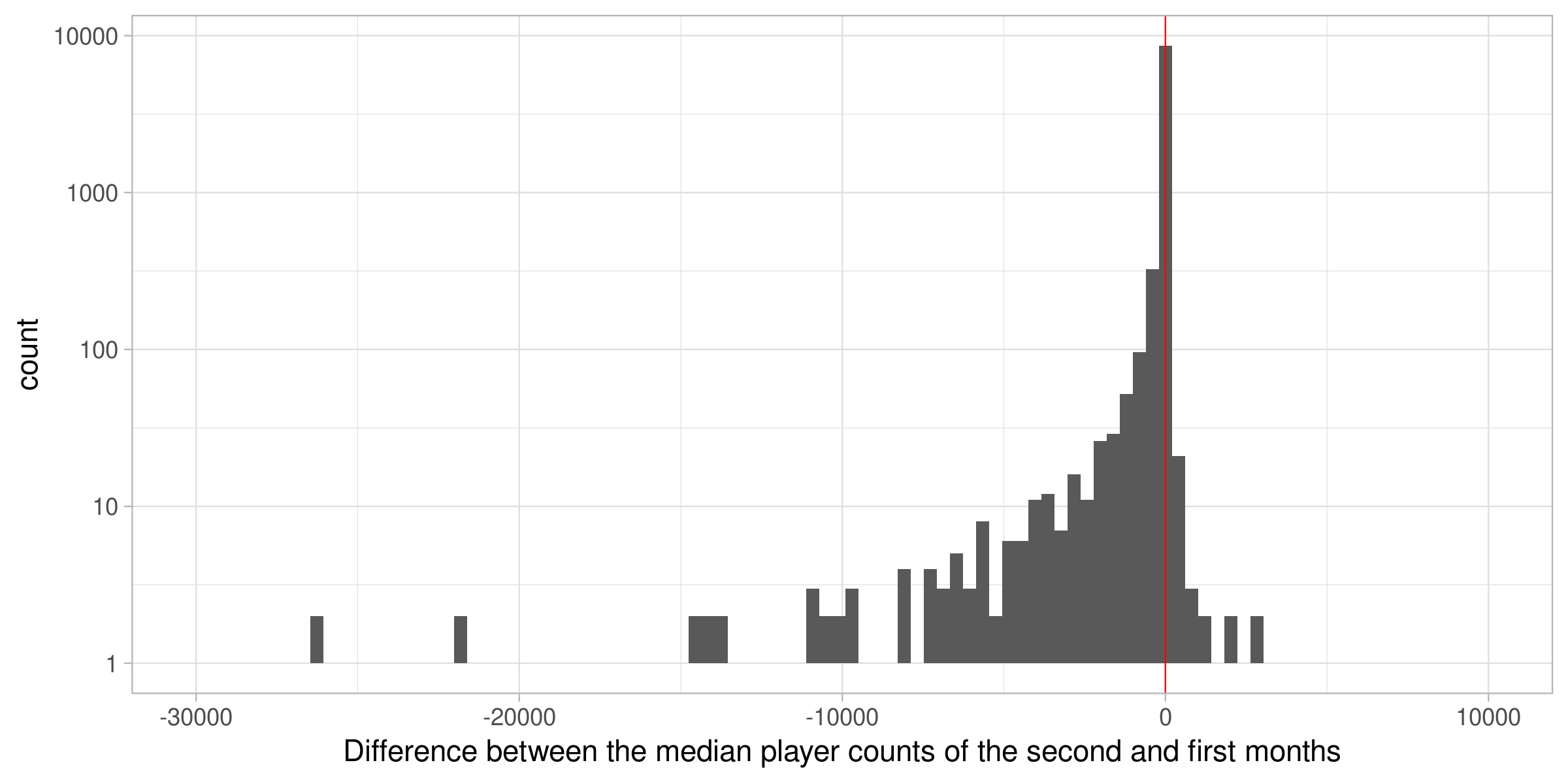}
    \caption{
        The difference between the main predictor and the target -- the medians of the second and first months. The y-axis is in log scale.
    }
    \label{fig:mediandiff}
\end{figure}

We are interested in the distribution of our target variable, the median count of players in the second month since a game's release. It relates heavily to our main predictor, the median player count of the first month. When visualized, both appear to be similar to a power law distribution, which is often the case with popularity in online media~\cite{ratkiewicz2010characterizing}. We can simplify the visualization by transforming the relevant data into just the difference between the player counts of the first and second months of the game -- more specifically, the difference between their medians. We show a histogram of the differences between the medians in \figurename~\ref{fig:mediandiff}. We can observe that most games do not see large changes in the number of players since most of the mass is near 0 and that most games lose players as the distribution leans more towards the left. The median of the differences is $-2$ while the mean is $-187$ as it is more affected by the outliers. We also generated equivalent plots for the third, fourth, and fifth month after the release, which all appear very similar to \figurename~\ref{fig:mediandiff}.

\begin{figure}
    \centering
    \includegraphics[width=\hsize]{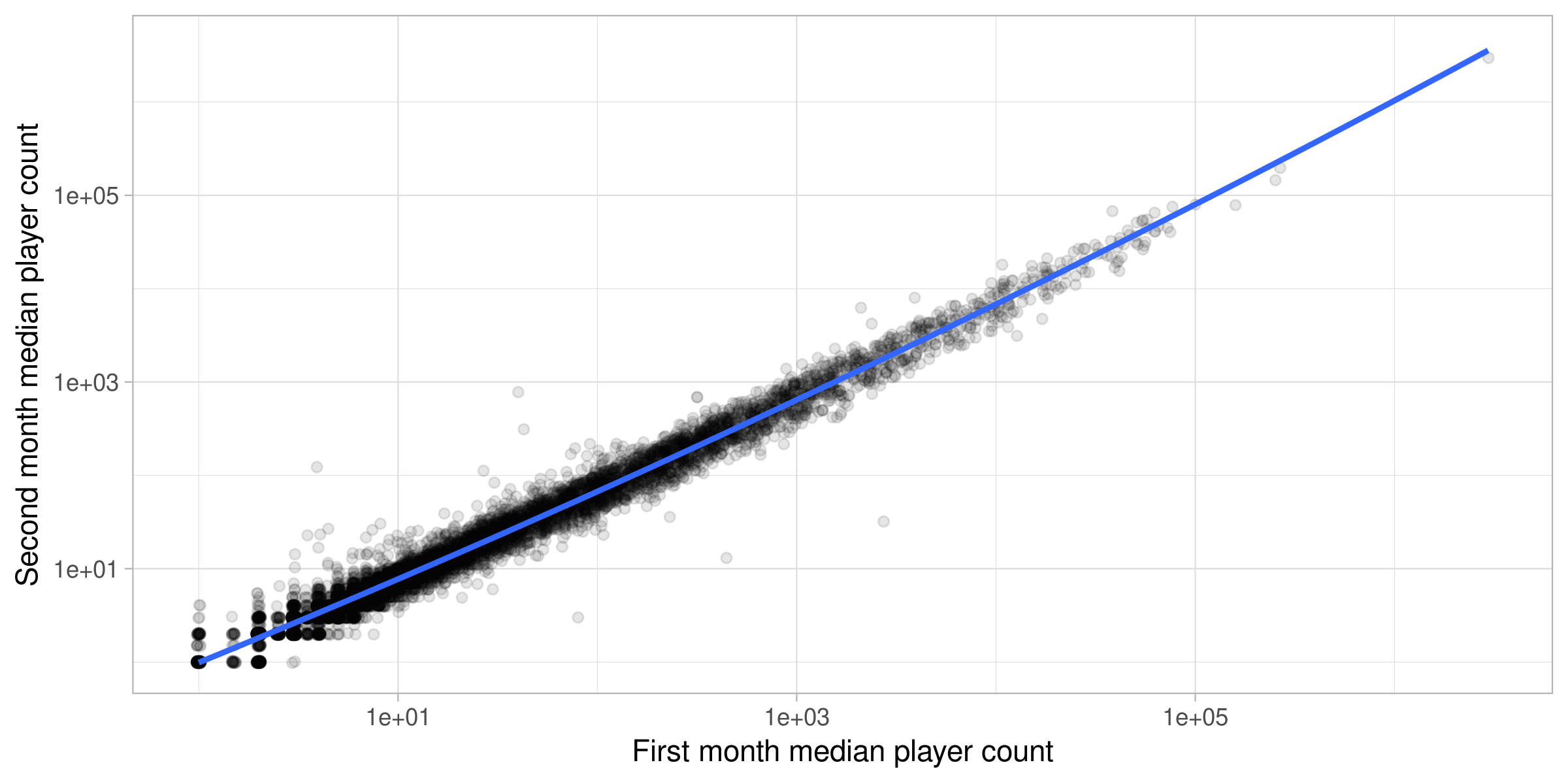}
    \caption{
        Correlation between the main explanatory variate and target -- medians of the second and first months.
        The blue line represents the trend while the dots are games. Both axes are in log scale.
    }
    \label{fig:mediancorr}
\end{figure}

We visualize the connection between the main predictor and the target (median player counts of the first two months) in \figurename~\ref{fig:mediancorr}. We can see that the trend is almost linear. The Pearson correlation coefficient between them is 0.998. This means that a model which just predicts the first month's median can be a decent baseline. In addition to the general linear correlation of median player counts, we also note the presence of some games that open with nearly 0 players in their first month and slowly climb as well as inverse behavior, where some games drop to 0 players in their second month. We also computed Pearson correlation coefficients for all other predictors and the target -- all of them are very low, from $-0.026$ to 0.011, meaning that there is very little linear correlation between them.
However, they may still affect the target in a non-linear manner.

\subsection{Feature transformations}

The features of past median player count, price, number of supported languages, and storage requirements were transformed with
\begin{equation}
\label{eq:log-transformation}
    f(x) = \log\Big( \frac{1 + x}{\overline{x}} \Big),
\end{equation}
where $x$ is the input feature and $\overline{x}$ is the mean of the feature in the training set. This approach is useful because our feature values are relatively small on average, but contain very large outliers (for example, the average value of \textit{past median player count} is roughly 1000, while its extremes range from 1 to 3,000,000). By adding 1 to our feature value, we avoid possible log-transformation problems in the following step. By dividing with the feature mean, we transform the raw feature value into a multiplier of the average observed value. Recalling the example above, this would transform our extreme values into 3000 and 0.001 multiplier values, respectively. Finally, log-transformation is performed to reduce the magnitude of our multipliers and make our features easier to model.

The value of \textit{release year} was treated as a categorical variable. Since outliers are not possible in the context of temporal features, we simply performed scaling, so both \textit{release day} and \textit{monthly release day} were moved to the $[0, 1]$ interval. Afterwards, we replaced both values with a tuple in the following manner:

$$x \mapsto [\sin(2 \pi x), \; \cos(2 \pi x)].$$.

This replacement is performed due to the cyclical nature of the temporal features. It ensures that games that were released at the beginning of the year and games which were released at the end of the year have similar release day values.

\subsection{Models}
Given the vector of the explanatory variates for each of the games, we aim to build a model predicting its median number of players in the second month after release. 
In particular, given the transformed features, we construct a final feature vector containing all attributes except for the genres, publishers, and developers. 
Given the final feature vector $\mathbf{x_i}$ of the $i$-th game, we would like to model its target number of players $y_i$.

\subsubsection{Normal model} \label{section:linear-regression}
We transformed the target in the same manner as we did for majority of the other features -- using the transformation in Equation~\ref{eq:log-transformation}. We modeled the transformed real-valued target as $$f(y) | \beta_0, \beta, \sigma^2 \sim \text{N}(\beta_0 + \mathbf{\beta}^T \mathbf{x}, \sigma^2),$$
$$\mathbf{\beta} \sim \text{Cauchy}(0, 5),$$
$$\sigma \sim \text{Half-Cauchy}(0, 5).$$
For $\mathbf{\beta}$, we introduced $\text{Cauchy}(0, 5)$ as weakly-informative and zero-centered prior due to our lack of knowledge in both positive or negative feature effect, and as a wider-tailed distribution allowing for less penalization of larger parameters. The uncertainty in this model is assumed not to vary with different inputs and is modeled with input-invariant parameter $\sigma$ having a similar weakly-informative $\text{Half-Cauchy}(0, 5)$ prior.

\subsubsection{Folded normal model} \label{section:folded-normal}

We empirically found that the normal model is sometimes unstable.
As a possible solution, we decided not to transform the target as in Equation~\ref{eq:log-transformation} and instead kept it in its original form.
Using a normal distribution was no longer suitable, because it's support was over the entire real line, whereas our target was now non-negative.
To account for this, we sought a distribution that assigned nonzero density to all points in $[0, \infty)$.
Having nonzero density at 0 lets us consider games with no players.

The folded normal distribution is one such candidate.
It is based on the normal distribution, but only has support on the non-negative reals with positive density at $0$. It generalizes the half-normal distribution in allowing the point with the highest density to be different than $0$.

Given that (i) our transformed target measurements were non-negative, (ii) games with no players were possible, (iii) unpopular games were more probable, and (iv) the most probable prediction was not necessarily $0$ for every game in our dataset, we decided to incorporate the folded normal distribution to model our prediction uncertainty for each game. Formally, the probability density function (PDF) $f_{FN}$ of the folded normal distribution $\text{FN}(\mu, \sigma^2)$ parametrized with $\mu \in \mathbb{R}$ and $\sigma^2 \in \mathbb{R^+}$ is defined as:
\begin{align}
    f_{FN}(x | \mu, \sigma^2) =& \frac{1}{\sqrt{2 \pi \sigma^2}} \left( e^{-\frac{(x - \mu)^2}{2\sigma^2}} + e^{-\frac{(x + \mu)^2}{2\sigma^2}} \right) =  \\
    =& f_N(x | \mu, \sigma^2) + f_N(-x | \mu, \sigma^2),
\end{align}
where $f_N(x)$ is the PDF of the normal distribution and $x$ is an arbitrary non-negative real.
We transformed the location parameter with the near-linear function $h(x) = \log(1 + e^x)$ for better model convergence and we modeled the target $y$ as $$y | \beta_0, \mathbf{\beta}, \sigma^2 \sim \text{FN}(\log(1 + e^{\beta_0 + \mathbf{\beta}^T\mathbf{x}}), \sigma^2),$$
$$\mathbf{\beta} \sim \text{Cauchy}(0, 5),$$
$$\sigma \sim \text{Half-Cauchy}(0, 5).$$
We used the same priors on $\mathbf{\beta}$ and $\sigma$ as for the normal model, following the same reasoning.

During preliminary testing, we found that the varying order of magnitude of the target is problematic.
Targets range from zero to more than a million players, which may lead to convergence difficulties during model training.
As a solution, we replaced the target $y$ with $\log(1 + y)$.
This kept the target on the same interval, but made its distribution significantly less skewed.
Our model is thus trained with the transformed target\footnote{We say that $\log(1 + y)$ is distributed according to the folded normal, where $y$ is the player count.}, however we may still transform the target back to the original space for ease of interpretation.

\subsubsection{Hierarchical folded normal model} \label{section:hierarchical-folded-normal}

We previously discussed that the publisher and developer variables were unsuitable for the model, because of the large number of unique values and difficulties in meaningfully transforming them.
On the other hand, the genre variable was more manageable.
If a game belongs to some particular genres, then their general properties can be used to further improve predictions for the game.
We implemented this idea by adding a hierarchical component to the folded normal model.

We associated distinct $\beta_0$ coefficients to different genres. If $\beta_{0,j}$ denotes the corresponding coefficient of the genre $j$ and $\overline{\beta_{0,G_i}}=\frac{1}{|G_i|} \sum_{j\in G_i} \beta_{0,j}$ denotes the coefficient mean over genres in genre set $G_i$ of the game $i$, then the hierarchical model of the target attribute we proposed is given as
$$
y_i | \overline{\beta_{0, G_i}}, \mathbf{\beta}, \mathbf{x_i}, \sigma^2 \sim \text{FN}(\log(1 + e^{\overline{\beta_{0, G_i}} + \mathbf{\beta}^T\mathbf{x_i}}), \sigma^2),
$$
$$\beta_{0,j} | \beta_0, \sigma_{\beta, 0} \sim \text{N}(\beta_0, \sigma_{\beta, 0}),$$
$$\beta_0, \mathbf{\beta} \sim \text{Cauchy}(0, 1),$$
$$\sigma_{\beta, 0} \sim \text{Half-Cauchy}(0, 1),$$
$$\sigma \sim \text{Half-Cauchy}(0, 5).$$
We linked the individual genre-specific coefficients in a hierarchical structure by imposing $\beta_{0,j} | \beta_0, $ $\sigma_{\beta, 0} \sim \text{N}(\beta_0, \sigma_{\beta, 0}).$
Similarly as before, we imposed $\text{Cauchy}(0,1)$ priors on $\beta_0$ and $\beta$, and a $\text{Half-Cauchy}(0,1)$ prior on $\sigma_{\beta, 0}$.
Again, we use $\log(1 + y)$ as the target.

\subsubsection{Heteroscedastic models}
The predictive variance plays an important role, because it gives us an estimate of uncertainty in our predictions.
Having a shared variance after transforming the target tells us how much the predicted player count will vary when considering the order of magnitude.
For example, suppose we transform the target with $\log(1+y)$ and observe $\sigma^2 = 1$.
Now consider two games with 10 and $10^{4}$ predicted players each, and use base-10 logarithm for ease of understanding.
Roughly speaking, this particular variance implies that the predicted player count will vary somewhere between 1 and 100 for the first game.
The count will vary between $10^{3}$ and $10^{5}$ for the second game.
This lets us know how uncertain the model is relative to the magnitude of the prediction.

While this approach is generally useful, the problem is that a single scalar is not very flexible and may yield unreasonably high uncertainty estimates for some games with many players.
It is better to consider each game's features and use them to compute its particular variance.
This approach lets the model provide better results for individual games and is still flexible enough to learn a single shared scalar if needed.
We refer to models which make use of this idea as heteroscedastic.

\begin{figure}
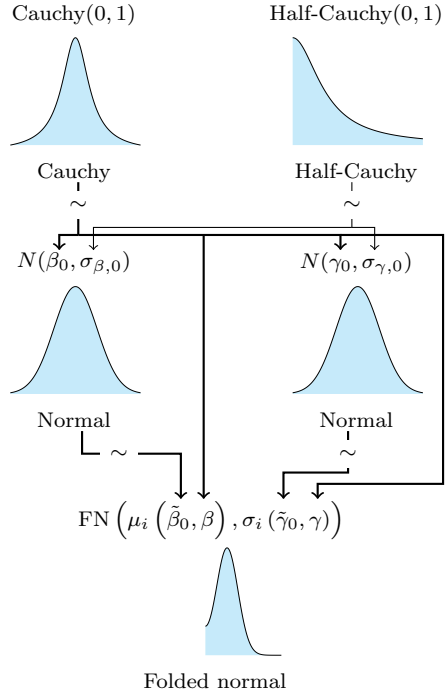

	\centering
	\begin{tikzpicture}
	[align=center,node distance=5mm]
	
	\node (md) at (0,0) {\scalebox{0.3}{{\input{img/tikz-files/folded-normal.tikz}}}};
	\node (mdtext) [below = -2.5mm of md] {\scriptsize Folded normal};
	\node (fndist) [above = -3mm of md] {\scriptsize $\text{FN}\left(\mu_{i}\left( \tilde{\beta}_{0}, \mathbf{\beta}\right), \sigma_{i}\left(\tilde{\gamma}_{0}, \mathbf{\gamma}\right) \right)$
	};

	\node (empty) [above = 2.3 of md] {};

	\node (n2) [right = of empty] {\scalebox{0.3}{\input{img/tikz-files/normal.tikz}}};
	\node (cctext) [below = -2.5mm of n2] {\scriptsize Normal};
	\node (cctext) [above = -2.5mm of n2] {\scriptsize $N(\gamma_0, \sigma_{\gamma, 0})$};
	
	\draw [->, thick] (1.75, 2.3) -- node[near end,fill=white] {\scriptsize$\sim$} (1.75, 1.9) --  (1.75, 1.75) -- (0.9, 1.75) -- (0.9, 1.4);
	
	\node (n1) [left = of empty] {\scalebox{0.3}{\input{img/tikz-files/normal.tikz}}};
	\node (cctext) [below = -2.5mm of n1] {\scriptsize Normal};
	\node (cctext) [above = -2.5mm of n1] {\scriptsize $N(\beta_0, \sigma_{\beta, 0})$};
	
	\draw [->, thick] (-1.75, 2.3) -- (-1.75, 2) -- node[near end,fill=white] {\scriptsize$\sim$} (-1.1, 2) -- (-0.45, 2) --  (-0.45, 1.4);
	
	\node (t1) [above = 5.6 of md] {};
	
	\node (tt1) [left = of t1] {\scalebox{0.3}{\input{img/tikz-files/cauchy.tikz}}};
	\node (cctext) [below = -2.5mm of tt1] {\scriptsize Cauchy};
	\node (cctext) [above = -2.5mm of tt1] {\scriptsize $\text{Cauchy}(0, 1)$};
	
	\draw [-, thick] (-1.8, 5.6) -- node[near end,fill=white] {\scriptsize$\sim$} (-1.8, 5.2) --  (-1.8, 4.9);
	\draw [->, thick] (-1.8, 4.9) -- (-2.05, 4.9) -- (-2.05, 4.7);
	\draw [->, thick] (-1.8, 4.9) -- (1.65, 4.9) -- (1.65, 4.7);
	\draw [->, thick] (-0.15, 4.9) -- (-0.15, 1.4);
	\draw [->, thick] (1.65, 4.9) -- (3, 4.9) -- (3, 1.6) -- (1.35, 1.6) -- (1.35, 1.4);
	
	\node (tt2) [right = of t1] {\scalebox{0.3}{\input{img/tikz-files/half-cauchy.tikz}}};
	\node (cctext) [below = -2.5mm of tt2] {\scriptsize Half-Cauchy};
	\node (cctext) [above = -2.5mm of tt2] {\scriptsize $\text{Half-Cauchy}(0, 1)$};
	
	\draw [-] (1.8, 5.6) -- node[near end,fill=white] {\scriptsize$\sim$} (1.8, 5.2) --  (1.8, 5);
	\draw [->] (1.8, 5) -- (-1.6, 5) -- (-1.6, 4.7);
	\draw [->] (1.8, 5) -- (2.1, 5) -- (2.1, 4.7);

\end{tikzpicture}
	\caption{Visualization of the heteroscedastic hierarchical folded normal model, where for conciseness we write $\mu_i(\tilde{\beta}_0, \beta) = \log\left(1 + e^{\frac{1}{|G_i|} \sum_{j\in G_i} \beta_{0,j} + \mathbf{\beta}^T\mathbf{x_i}}\right)$ and $\sigma_{i}(\tilde{\gamma}_0, \gamma) = e^{\frac{1}{|G_i|} \sum_{j\in G_i} \gamma_{0,j} + \mathbf{\gamma}^T\mathbf{x_i}}$ .}
	\label{fig:tikzmodel}
\end{figure}

We thus replaced the $\sigma^2$ parameter of all three models with a function of a game's features.
More precisely, we used $e^{\gamma_0 + \mathbf{\gamma}^T \mathbf{x}}$ for the normal and the folded normal model, and $e^{\overline{\gamma_{0, G_i}} + \mathbf{\gamma}^T\mathbf{x_i}}$ for the hierarchical normal model with $\gamma_{0,j} | \gamma_0, \sigma_{\gamma, 0} \sim N(\gamma_0, \sigma_{\gamma, 0})$. 
This approach yielded three new models with attribute-dependent variance.
The priors for $\gamma$ coefficients were the same as those for $\beta$ coefficients in the homoscedastic models.
The heteroscedastic hierarchical folded normal model is the most complex of the three.
We visualize it in Figure~\ref{fig:tikzmodel}. 

\section{Results}\label{sec:results}

\subsection{Visual check of the predictive distributions}

\begin{figure}
    \centering
    \includegraphics[width=1.0\hsize]{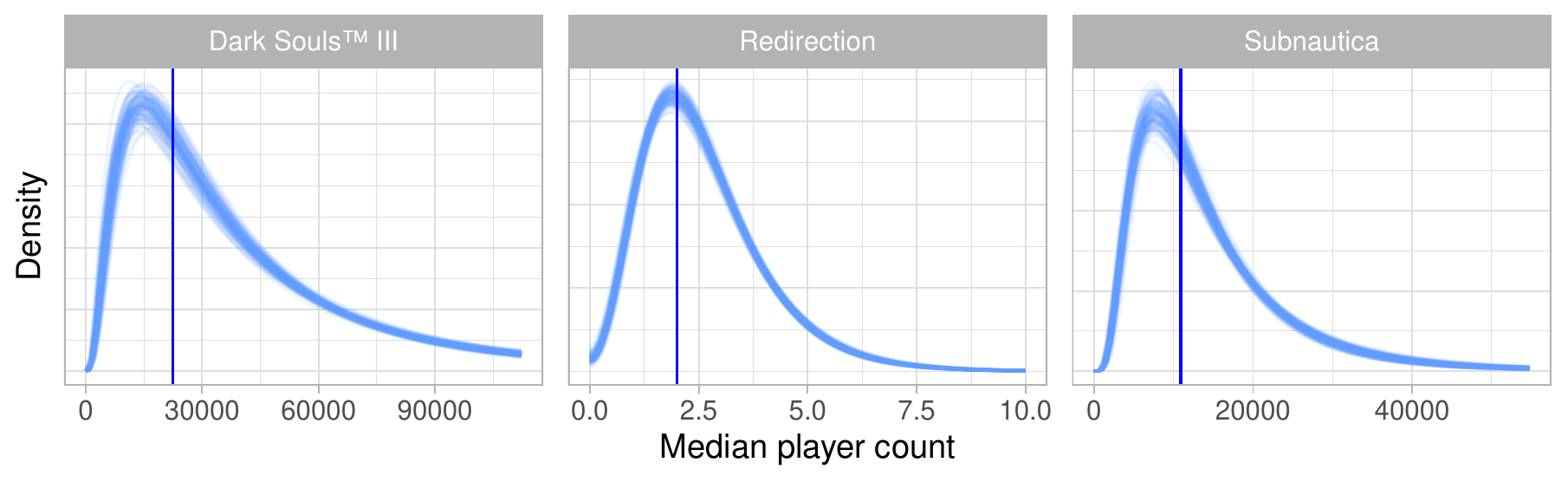}
    \caption{
        Predicted median player count distributions using the Hierarchical folded normal.
        Vertical lines represent the targets.
        Dark Souls\textsuperscript{TM} has the highest overall budget, followed by Subnautica, and Redirection.
        Based on these examples, we see that the model can perform well for games with such differences.
    }
    \label{fig:single-game-distributions}
\end{figure}

By sampling from the posterior, we compute different game-specific parameters $\mu$ and $\sigma$ which correspond to some distribution that describes the median player count for that game.
We visualize some predictions with the heteroscedastic hierarchical folded normal model in Figure~\ref{fig:single-game-distributions}.

\subsection{Model comparison}

We evaluate and compare the models' performances with an approximation of the leave-one-out cross-validation (LOOCV) technique, the Pareto smoothed Importance sampling Leave-one-out information criterion (PSIS-LOOIC), first presented by Vehtari et al. (2016) \cite{Vehtari_2016}. We decided to use the LOOCV approach instead of a simple holdout estimation to get more accurate results, but LOOCV is an exhaustive model evaluation technique and it requires refitting the model once per each data instance (becoming infeasible on big datasets like ours). Therefore we approximated it with LOOIC -- a fast, robust, and stable model evaluation method, based on the log-likelihood of the posterior at the actual target, and designed specifically for Bayesian models.

\begin{figure}[ht]
    \centering
    \includegraphics[width=\hsize]{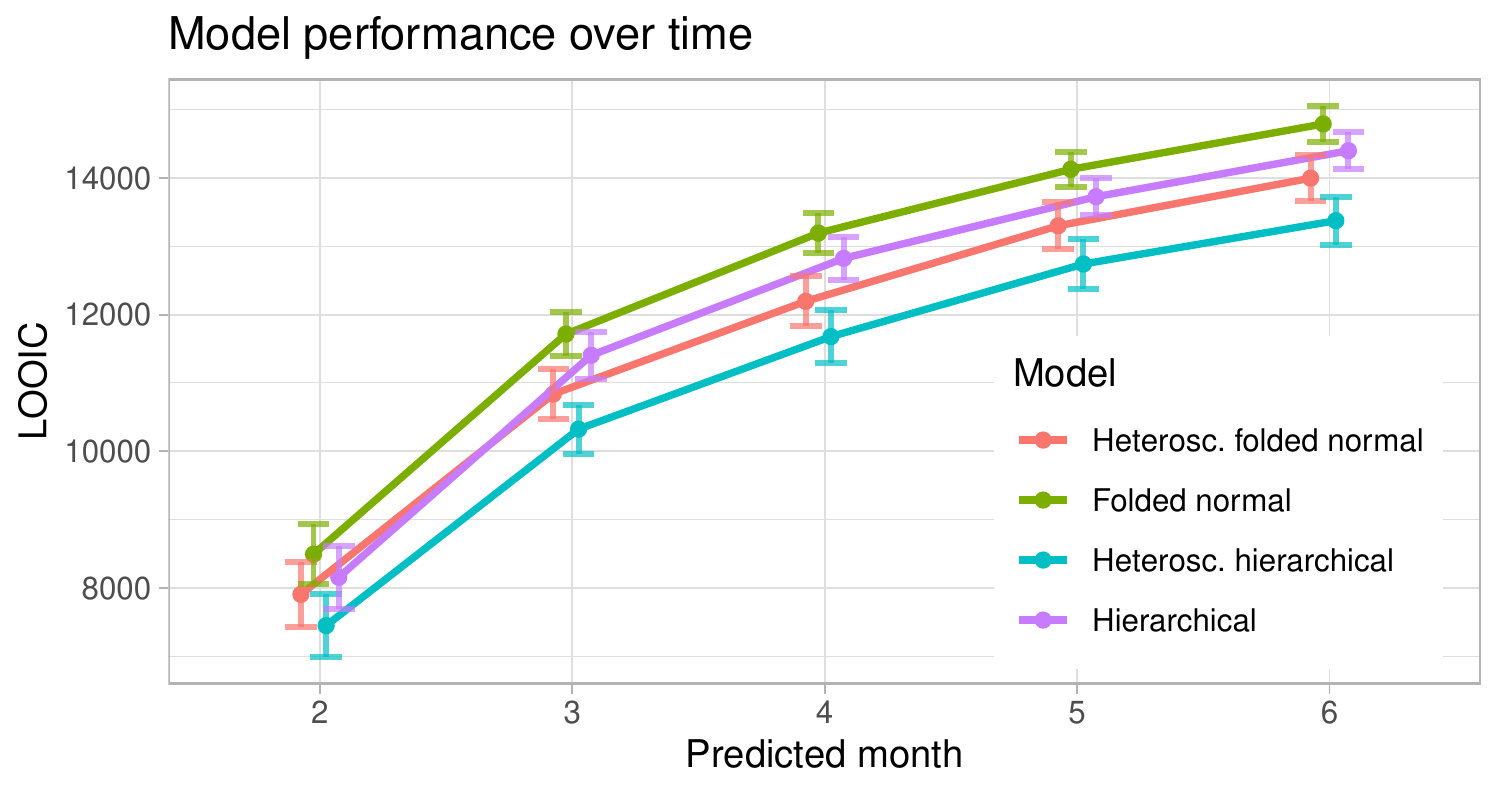}
    \caption{LOOIC estimated for multiple predicted months. For each model-predicted month pair, we show the LOOIC estimate and its standard error. The performance of the models expectedly drops over time, with the greatest decrease between the 2nd and 3rd month (notice that the lower the LOOIC estimate is the better is the model performing). We include multiple models on this plot to give the reader a feeling about their performances, but any conclusions about which one is better should not be made from it.}
    \label{fig:looic_time}
\end{figure}

\begin{figure}[ht]
    \centering
    \includegraphics[width=1\textwidth]{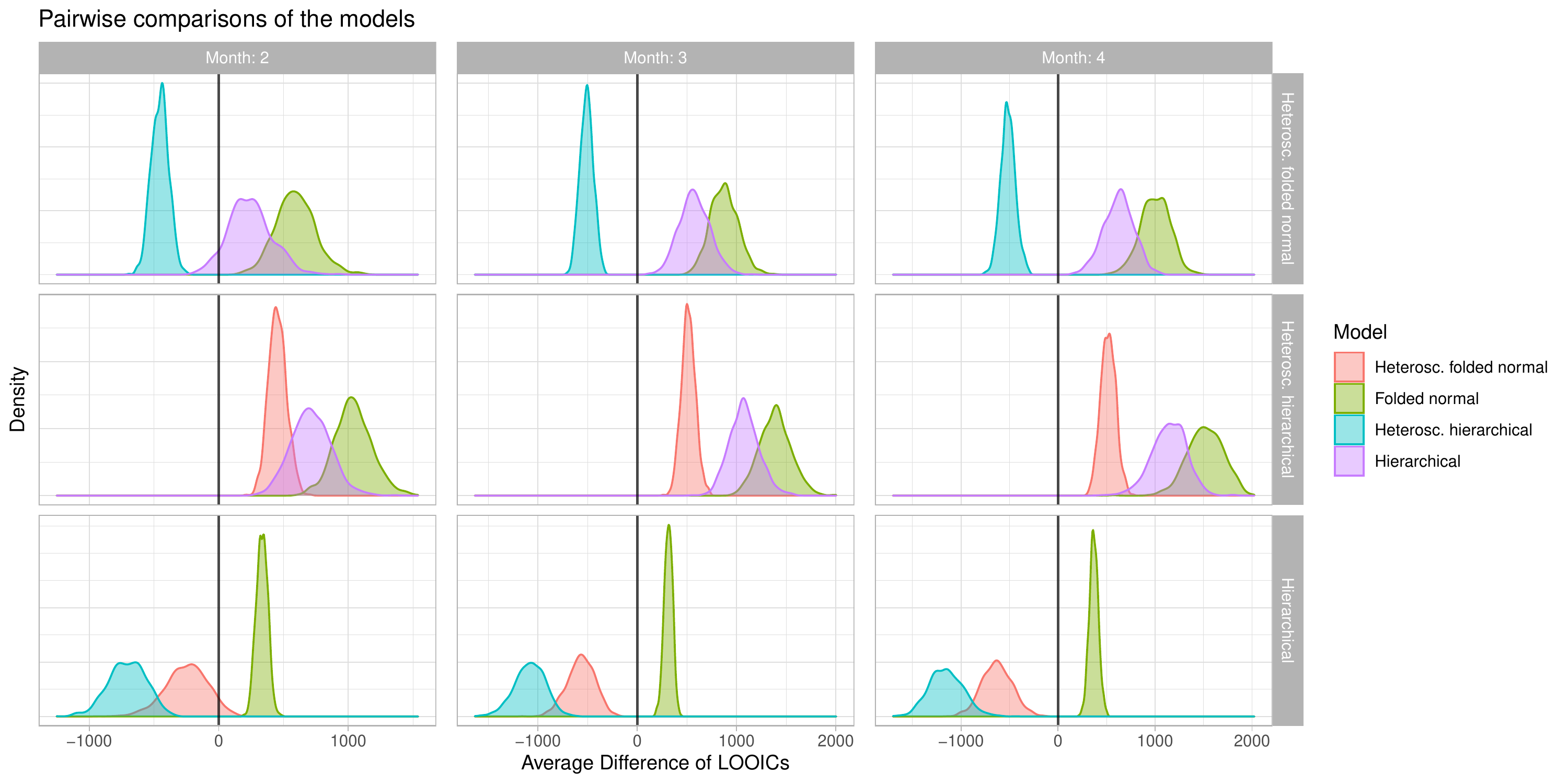}
    \caption{Distributions of LOOIC differences between the models. We compute the differences by subtracting the pointwise LOOICs estimations of the compared models. We obtain the distributions by bootstrapping the computed differences. Each model should be compared only with that line's reference model (multiple models are shown simultaneously just for convenience). For a selected model and the reference to which we want to compare it, the percentage of the area under its density curve that lies on the left of the $x=0$ line is the confidence with which we can claim the selected model is better than the reference. For example, from the visualization, we can conclude that no matter the predicted month, the heteroscedastic hierarchical model is almost surely the best one. Furthermore, we can be almost sure of the difference between the heteroscedastic folded normal's and heteroscedastic hierarchical models' LOOIC being at least 200.}
    \label{fig:pairwise_comp}
\end{figure}

For each data instance, LOOIC approximates the expected log-predictive density (\emph{elpd}) of the model fitted on the dataset from which the selected game would be removed.\footnote{PSIS-LOOIC is thoroughly described by Vehtari et al. (2016)~\cite{Vehtari_2016} and Vehtari et al.~(2002) \cite{Vehtari_2002}.} The LOOIC value is then computed as $\text{LOOIC} = -2 \cdot elpd$ to get on the deviance scale.
The principal advantage that LOOIC provides is its low time complexity. While LOOCV requires refitting the model N times (with N the size of the dataset), LOOIC requires only one evaluation of the model. 
We show the models' performances over time in Figure \ref{fig:looic_time}. All the models exhibit a similar (and expected) behavior, with their performance dropping when predicting player counts further in the future. We present a proper pairwise comparison between the models in Figure \ref{fig:pairwise_comp}. The heteroscedastic hierarchical model (which is the most complex one) turns out to be the best performing model. The other models might provide viable alternatives if a faster fitting process is desired.

\subsection{Coefficient interpretation}
We analyze the posterior distributions of sampled coefficients.
Because different features use somewhat different scales, we do not claim that any feature is more important in predicting the outcome than another.\footnote{We further analyzed these features to see if they could be transformed to a common distribution, which would make the posteriors of the coefficients easier to compare. Since no pair of features could meaningfully be transformed to distributions of similar shapes, we concluded that pairwise comparisons would likely be unreliable and did not pursue this any further.}
However, we do quantify the contribution of each feature towards the computed parameters of the folded normal distribution.
We focus on the heteroscedastic folded normal model.
The posterior for this model is presented in Table~\ref{tab:folded-normal-posterior}.
Since the hierarchical variant of the model has a very similar posterior, we only explicitly analyze the genre-specific intercepts.

\begin{table}[]
    \begin{tabular}{l}
        \textrm{Feature} \\ \hline
        \textrm{Intercept} \\
        \textrm{Price} \\
        \textrm{Number of languages} \\
        \textrm{Storage requirements} \\
        \textrm{Median players feature} \\
        \textrm{Day of month} \\
        \textrm{Day of year, cosine} \\
        \textrm{Day of year, sine}
    \end{tabular}
    \quad
    \begin{tabular}{l|rrr}
                              & $\mu$    & $q_5$    & $q_{95}$   \\ \hline
        $\beta_0$ & $7.583$ & $7.539$ & $7.627$ \\
        $\beta_1$ & $-0.001$ & $-0.002$ & $0.000$ \\
        $\beta_2$ & $-0.039$ & $-0.049$ & $-0.029$ \\
        $\beta_3$ & $0.013$  & $0.000$ & $0.027$  \\
        $\beta_4$ & $0.988$  & $0.982$  & $0.994$  \\
        $\beta_5$ & $-0.022$ & $-0.030$ & $-0.016$ \\
        $\beta_6$ & $0.055$  & $0.041$  & $0.070$  \\
        $\beta_7$ & $-0.005$ & $-0.018$ & $0.010$ 
    \end{tabular}
    \quad
    \begin{tabular}{l|rrr}
          & $\mu$    & $q_5$    & $q_{95}$   \\ \hline
    $\gamma_0$ & $-0.259$ & $-0.316$ & $-0.203$ \\
    $\gamma_1$ & $-0.002$ & $-0.003$ & $0.000$ \\
    $\gamma_2$ & $-0.068$ & $-0.082$ & $-0.054$ \\
    $\gamma_3$ & $-0.026$  & $-0.044$ & $-0.007$  \\
    $\gamma_4$ & $0.102$  & $0.094$  & $0.110$  \\
    $\gamma_5$ & $-0.060$ & $-0.070$ & $-0.051$ \\
    $\gamma_6$ & $0.057$  & $0.037$  & $0.076$  \\
    $\gamma_7$ & $0.068$ & $0.049$ & $0.088$
    \end{tabular}
    \caption{Coefficients to compute the mean (using $\beta_i$) and variance (using $\gamma_i$) of the folded normal distribution. The columns represent the posterior mean, 5th, and 95th percentiles. The 90\% confidence intervals for price, storage requirements, and day of year (sine component) include 0, so their contribution to the mean and the variance is probably negligible.}
    \label{tab:folded-normal-posterior}
\end{table}

First, consider the main non-negative features -- price, number of supported languages, system requirements, and median player count.
Recall that the folded normal model and its hierarchical variant use the exponential of the dot product to compute the transformation-specific parameter (the mean and the variance).
We use an intercept term as a reference, so each coefficient individually affects the dot product in an additive manner.
Since the four considered features are all transformed using $\log\left( \frac{1 + x}{\overline{x}} \right) = \log(1+x) - \log(\overline{x})$, a coefficient $\beta_i$ causes one of the following changes in the computed parameter:

\begin{itemize}
    \item if $\beta_i < 0$, the dot product value is decreased by $|\beta_i|\log(1+x)$, then increased by $|\beta_i|\log(\overline{x})$;
    \item if $\beta_i > 0$, the dot product value is increased by $|\beta_i|\log(1+x)$, then decreased by $|\beta_i|\log(\overline{x})$;
    \item if $\beta_i = 0$, then there is no change in the dot product value.
\end{itemize}

\begin{figure}
    \centering
    \includegraphics[width=\hsize]{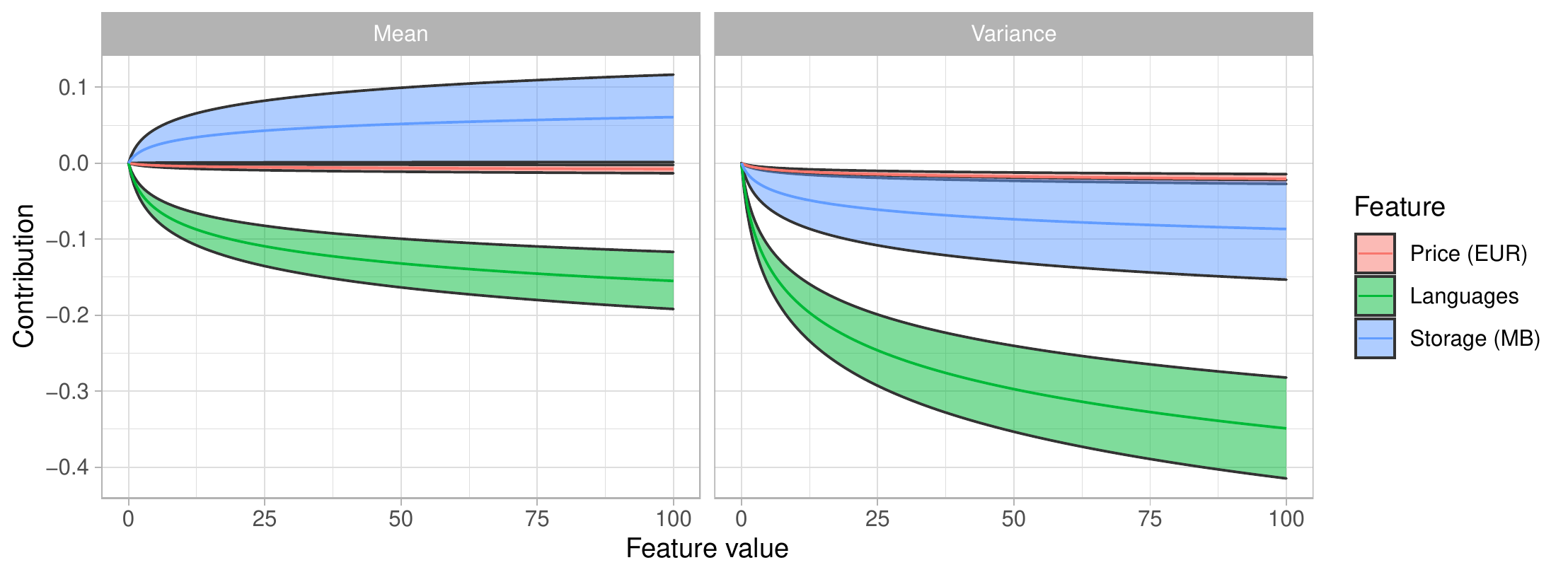}
    \caption{
        Parameter contribution plot for three of the four main features.
        The median player count feature is omitted for easier visualization.
        Parameter contribution is computed by multiplying the parameter with $x$, which is a feature in the original space, i.e.\ before the $\log(1+x)$ transformation.
        Each feature is represented by its mean contribution and the 90\% confidence interval across different values.
        The left plot shows that as the number of languages increases, the dot product and thus the mean of the folded normal are decreased.
        On the other hand, increasing storage is associated with an increase in the dot product and consequently the mean.
        On the right plot, we see that all three features cause a drop in variance.
        The price feature has an almost negligible effect on the computed parameters.
    }
    \label{fig:feature-contributions}
\end{figure}

Note that $\overline{x} > 1$ for all four features, so each $\log(\overline{x})$ term is strictly positive.
We can treat all such terms as constants and focus only on $|\beta_i|\log(1 + x)$, which are different across games.\footnote{
    If we explicitly account for the four feature means $\overline{x}$ by subtracting them from the base intercept, we arrive at an adjusted intercept $\widetilde{\beta_0}$ with mean $1.443$ and 90\% confidence interval (CI) $(1.326, 1.555)$.
    The corresponding result for the variance is $\widetilde{\gamma_0}$ with mean $-0.670$ and 90\% CI $(-0.714, -0.425)$.}
We visualize their contributions in Figure~\ref{fig:feature-contributions}.
The median player count feature is associated with an increase in the mean and variance parameters, but is omitted for a clearer visualization.
Its contribution is larger than that of the other three features, which is evident from the previously shown side-by-side correlation plot with the target (Figure~\ref{fig:mediancorr}).

\begin{figure}
    \centering
    \includegraphics[width=\hsize]{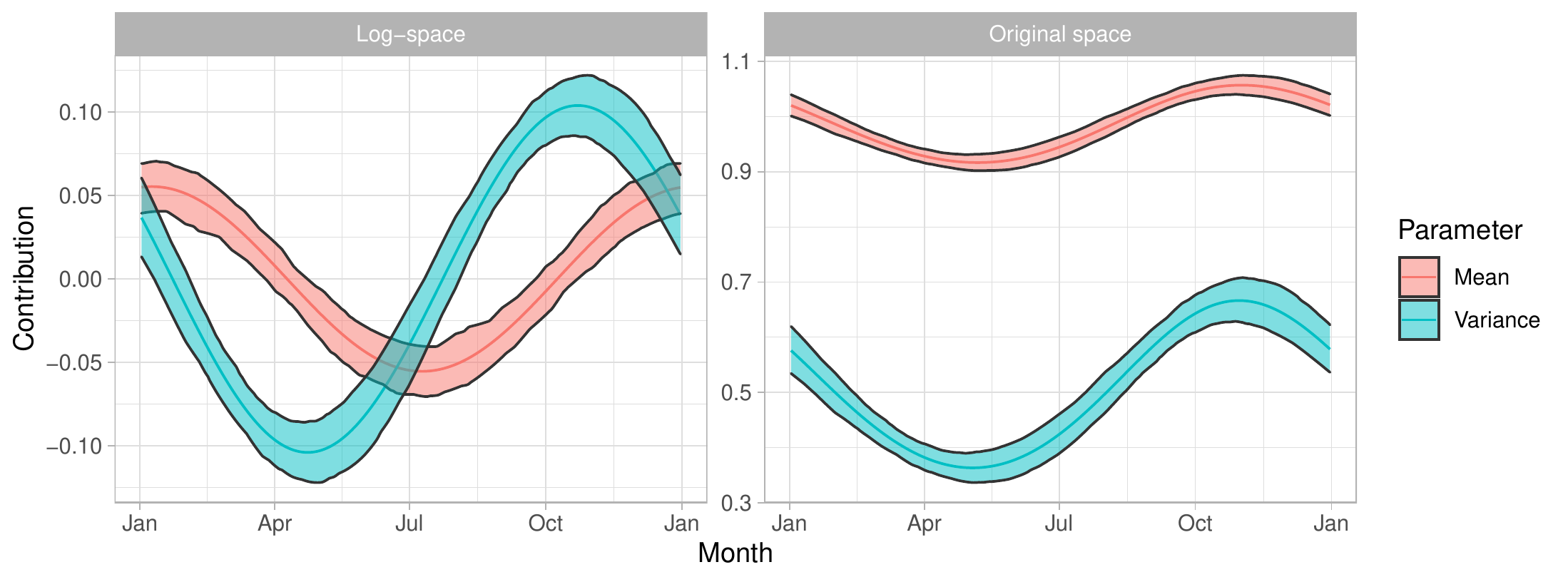}
    \caption{
        Parameter contribution plot for the \textit{release day} feature.
        The plot on the left represents parameter contribution in log-space, whereas the plot on the right is its transformation to the original space with actual player counts.
        Releasing a game around January is associated with the highest increase in the mean parameter, whereas releasing it around July is associated with the highest decrease.
        In log-space (i.e.\ referring to the folded normal parameters) the variance parameter follows a similar pattern, shifted a few months backwards.
        In the original space (i.e.\ considering actual player counts), the mean and the variance are aligned.
        We can see that the highest contribution roughly between October and January, and the lowest roughly between May and August.
        On a related note, seasonal behavior has previously been observed for movie releases~\cite{radas1998seasonal}, where peak release counts occurred before holidays.
        Our data includes total player counts from all countries where Steam is available, so national holidays do not present a significant contribution.
        A possible explanation for the peak is that publishers aim to release games before the globally celebrated New Year's holidays, so the games are available during large sales and thus gain more visibility.
    }
    \label{fig:date-contribution}
\end{figure}

The temporal features also have an effect on the computed parameters.
In a similar manner as before, we inspect the contribution that the \textit{release day} feature has on the computed parameters in Figure~\ref{fig:date-contribution}.

The posterior for the \textit{monthly release day} coefficient is negative for the mean, which suggests that releasing a game later in the month decreases the number of players.
According to the model, releasing the game on the 31st day of a month is associated with an expected decrease of the folded normal mean by $-0.682$, which is non-negligible considering the contributions of other features.
We did not find a clear explanation for this phenomenon, however we still use the feature as it improves model predictions.

\begin{figure}%
    \centering
    \subfloat[\centering Mean intercepts]{{
        \includegraphics[width=0.45\hsize, trim={0 0 0 12mm}, clip]{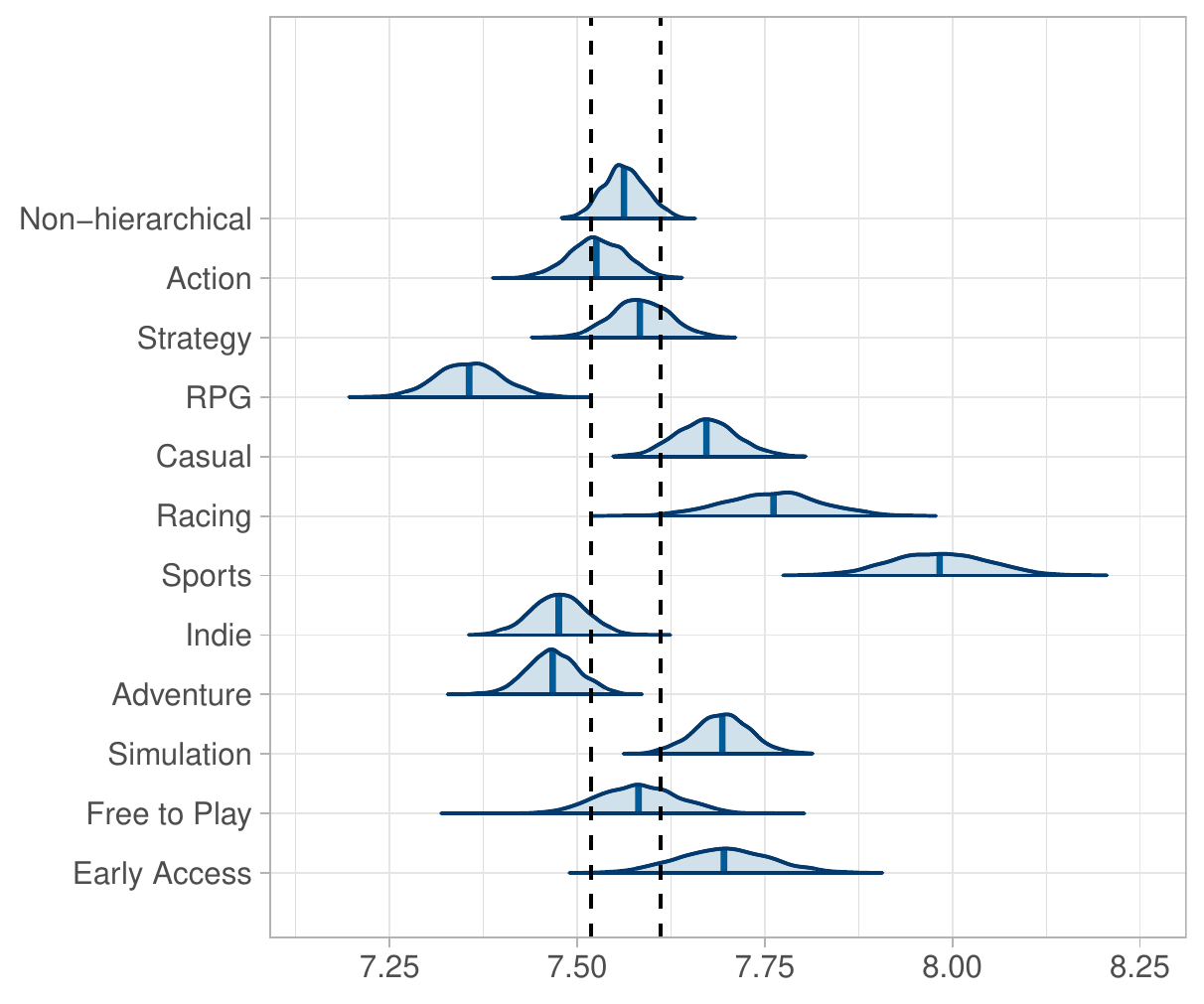} 
    }}%
    \qquad
    \subfloat[\centering Variance intercepts]{{
        \includegraphics[width=0.45\hsize, trim={0 0 0 12mm}, clip]{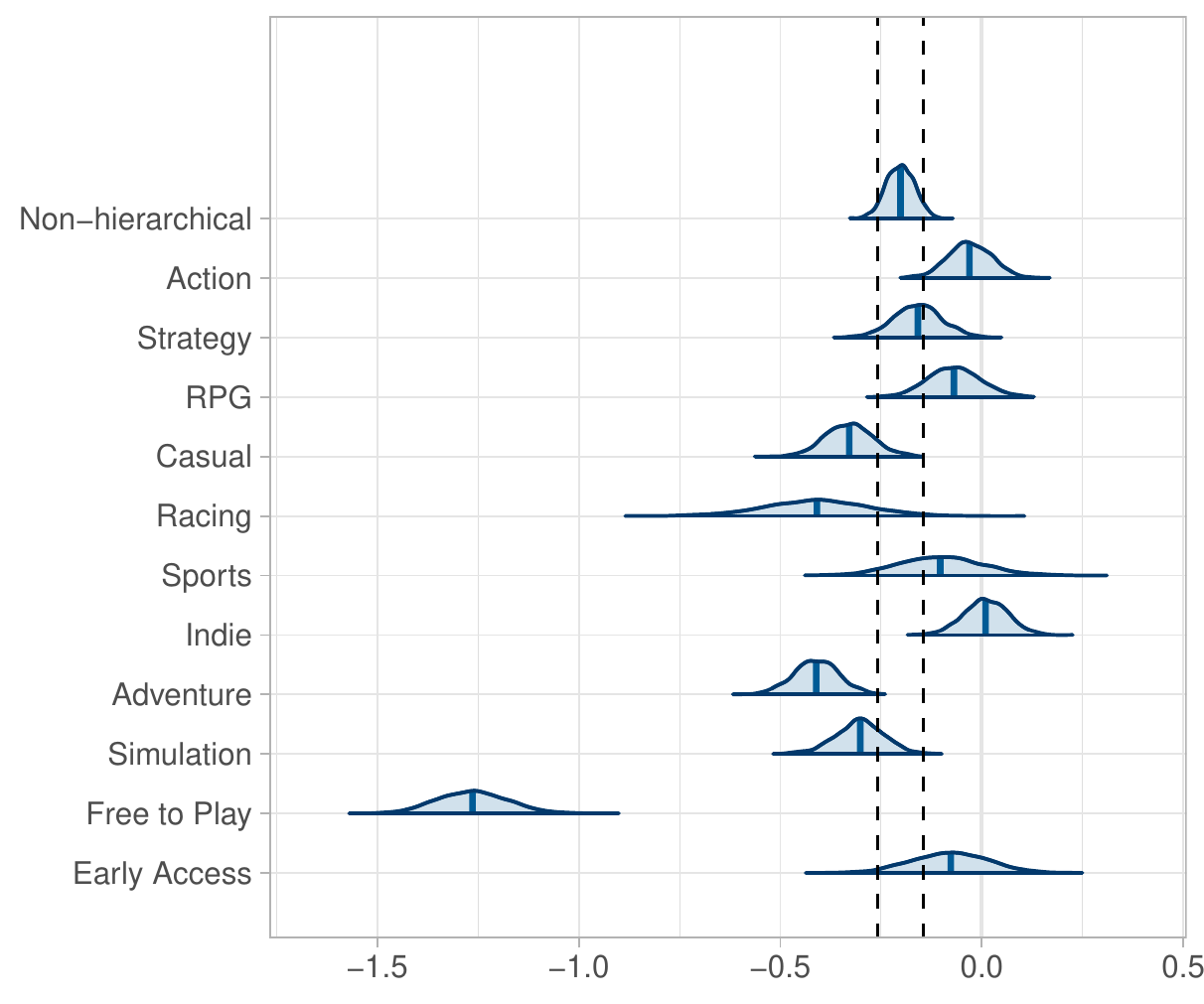}
    }}%
    \caption{
        Visualization of the genre-specific intercepts for computing the mean (left) and variance (right) of the heteroscedastic hierarchical folded normal model.
        Dashed lines represent the $[q_{5}, q_{95}]$ interval for the \textit{non-hierarchical} heteroscedastic folded normal posterior.
        Some genres were skipped due to high-variance posteriors.
        The left plot indicates that RPG and indie games tend to a smaller intercept, whereas sports, racing, simulation, and possibly casual games attain a higher intercept.
        In the right plot, we see that most genres are similar to the regular model.
        The exceptions are free to play games, which have a significantly smaller variance and thus result in more stable predictions.
        The predictions for adventure games are also more stable with this model, whereas the variance of predictions for indie games is somewhat greater.
        In conclusion, the heteroscedastic hierarchical model estimates the folded normal mean more precisely for these genres, because their distributions are clearly different from the regular model and also achieve a reasonably small variance.
        The model also helps by reducing variance in free to play game predictions.
        From a simple visual check, the variance of other genres revolves around the regular model's variance and does negatively affect predictive performance by comparison.
    }
    \label{fig:hierarchical-intercepts}
\end{figure}

The heteroscedastic folded normal uses individual $\beta_0$ and $\gamma_0$ intercepts to compute its mean and variance parameters.
On the other hand, the hierarchical variant considers $\beta_0$ and $\gamma_0$ parameters for each genre individually, then computes the average over all of the game's genres.
Each genre thus influences predictions to a different degree.
A visualization of these coefficients is presented in Figure~\ref{fig:hierarchical-intercepts}.

\begin{figure}
    \centering
    \includegraphics[width=1.0\hsize]{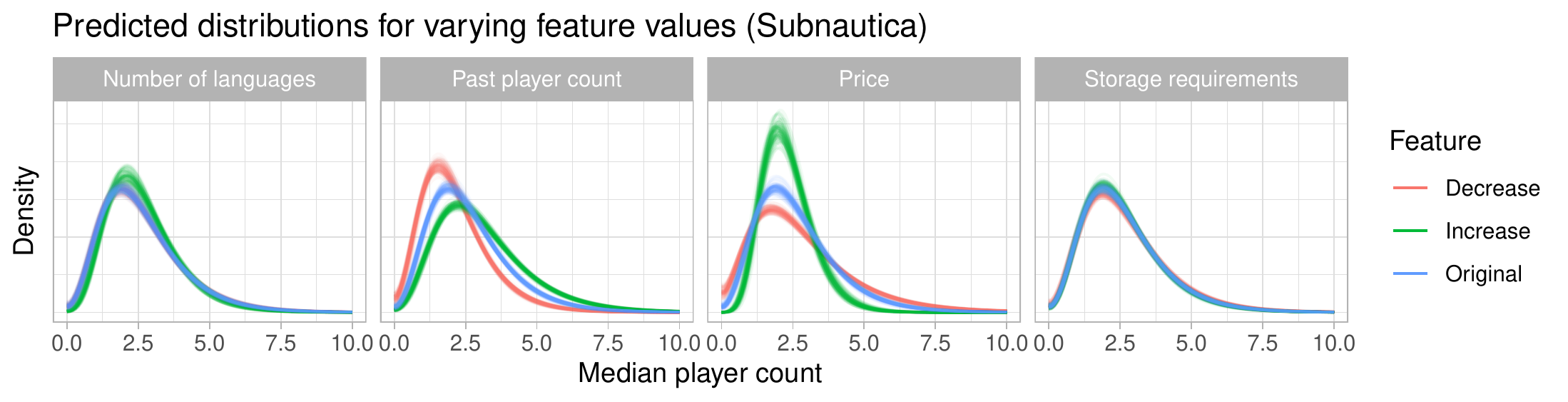}
    \caption{
    Intuition of how changing feature values affects predicted distributions for the game Subnautica.
    We use the heteroscedastic hierarchical folded normal model for predictions.
    Listed as triples, the decreased, original, and increased feature values are (1, 18, 29) for the number of languages; (20000, 27143, 40000) for past median player count; (0 EUR, 21.2 EUR, 60 EUR) for price; (10 GB, 20 GB, 100 GB) for storage requirements.
    In this example, changing the number of languages and the storage has a very small effect on the generated distributions.
    Changing the price results in noticeably different predictions.
    As established before, the target variable is heavily correlated with the past median player count feature, so the changes in generated distributions are the largest here.
    }
    \label{fig:feature-changes}
\end{figure}

Since we did not use a GLM as in its formal definition, we cannot easily state how model predictions behave as we change some input features.
To provide some intuition, we visualize such changes for a particular game in Figure~\ref{fig:feature-changes}.

\section{Conclusion}\label{sec:conclusion}

We constructed and evaluated models to predict the popularity of games on Steam in the early stages after their release while also analyzing the effect of other features. We represented game popularity with the number of players playing a game over time. 

We manually collected the game data from different sources and constructed various features through different transformations. Our main prediction target was the median player count in the second month after a game's release. We also experimented with predicting further in time and found that the models' performance decreased. Since understanding the effect of the features was our main goal, we focused mostly on the second month target which was the easiest to model. We also noticed that models that did not use the median player count of the first month as a feature did not converge. This likely occurs because the model predicts player counts in absolute terms and requires a good starting point, which is then shifted to obtain the prediction. This issue could be resolved by using additional features to identify the starting point or by reformulating the model to predict relative changes in player count as percentages instead of absolute values. It is not guaranteed that the second approach would alleviate all such problems, because some games would still show unexpectedly quick changes in the number of players, but it could serve as a good starting point.

We evaluated Bayesian normal, folded normal, and hierarchical folded normal models in both homo- and heteroscedastic variants for this task. We augmented the baseline, the first month's median player count with other features and then examined their influence with the help of their posterior distributions. The heteroscedastic hierarchical model achieved the best performance with respect to LOOIC. The most important predictor in all models was the median player count, which makes sense as it is also a strong baseline. However, we found that adding additional features benefitted the models and enabled us to analyze and interpret their effects.

We see great potential for future work in this area, especially in the analysis parts. Gathering and adding more game features could improve the models substantially -- this will likely get easier with time as Steam will gather and offer more data. 

Another similar improvement would be to perform a thorough analysis of missing values of player counts and incorporate them into the models. A more complex and difficult goal would be to model the player counts from and to any arbitrary pair of points in time. This would enable a much deeper level of understanding how the popularity changes over time.

A possible enhancement that would likely improve this modeling approach would be adding other popularity-related features such as Twitch views, Google searches, reviews, etc. To fully utilize reviews, it is also possible to use natural language processing to extract certain features. Seeing that predicting a game's popularity accurately would be very valuable for publishers or even streamers, we see a possible practical application for predictive models in this area.

\thispagestyle{empty}
\scriptsize
\bibliographystyle{unsrt}
\bibliography{references.bib}

\begin{thebibliography}{10}

\bibitem{budiarto2018game}
Joseph~Alexander Budiarto.
\newblock Game popularity tracking system.
\newblock {\em International Journal of Industrial Research and Applied
  Engineering}, 3(2):79--85, 2018.

\bibitem{ahn2017makes}
Sangho Ahn, Juyoung Kang, and Sangun Park.
\newblock What makes the difference between popular games and unpopular games?
  analysis of online game reviews from steam platform using word2vec and bass
  model.
\newblock {\em ICIC Express Letters}, 11(12):1729--1737, 2017.

\bibitem{lin2019empirical}
Dayi Lin, Cor-Paul Bezemer, Ying Zou, and Ahmed~E Hassan.
\newblock An empirical study of game reviews on the steam platform.
\newblock {\em Empirical Software Engineering}, 24(1):170--207, 2019.

\bibitem{lin2018empirical}
Dayi Lin, Cor-Paul Bezemer, and Ahmed~E Hassan.
\newblock An empirical study of early access games on the steam platform.
\newblock {\em Empirical Software Engineering}, 23(2):771--799, 2018.

\bibitem{steamspysite}
Steam{S}py.
\newblock \url{https://steamspy.com}.
\newblock Accessed: 2020-12-17.

\bibitem{steamdbsite}
Steam{DB}.
\newblock \url{https://steamdb.info}.
\newblock Accessed: 2020-12-18.

\bibitem{liu_2010}
Hongju Liu.
\newblock Dynamics of pricing in the video game console market: Skimming or
  penetration?
\newblock {\em Journal of Marketing Research - J MARKET RES-CHICAGO},
  47:428--443, 06 2010.

\bibitem{limenet_2020}
Market research: The state of online gaming – 2020.
\newblock
  \url{https://www.limelight.com/resources/white-paper/state-of-online-gaming-2020/}.
\newblock Accessed: 2021-02-21.

\bibitem{palomba_2019}
Anthony Palomba.
\newblock Digital seasons: How time of the year may shift video game play
  habits.
\newblock {\em Entertainment Computing}, 30:100296, 03 2019.

\bibitem{ratkiewicz2010characterizing}
Jacob Ratkiewicz, Santo Fortunato, Alessandro Flammini, Filippo Menczer, and
  Alessandro Vespignani.
\newblock Characterizing and modeling the dynamics of online popularity.
\newblock {\em Physical review letters}, 105(15):158701, 2010.

\bibitem{Vehtari_2016}
Aki Vehtari, Andrew Gelman, and Jonah Gabry.
\newblock Practical bayesian model evaluation using leave-one-out
  cross-validation and waic.
\newblock {\em Statistics and Computing}, 27(5):1413–1432, Aug 2016.

\bibitem{Vehtari_2002}
Aki Vehtari and Jouko Lampinen.
\newblock Bayesian model assessment and comparison using cross-validation
  predictive densities.
\newblock {\em Neural computation}, 14:2439--68, 11 2002.

\bibitem{radas1998seasonal}
Sonja Radas and Steven~M Shugan.
\newblock Seasonal marketing and timing new product introductions.
\newblock {\em Journal of Marketing Research}, 35(3):296--315, 1998.

\end{thebibliography}
	
\end{document}